\documentclass{article} 
\usepackage{iclr2021_conference,times}

\usepackage{amsmath,amsfonts,bm}

\def\eqref#1{equation~\ref{#1}}

\def\1{\bm{1}}

\DeclareMathAlphabet{\mathsfit}{\encodingdefault}{\sfdefault}{m}{sl}
\SetMathAlphabet{\mathsfit}{bold}{\encodingdefault}{\sfdefault}{bx}{n}

\usepackage{hyperref}
\usepackage{url}
\usepackage{graphicx}
\usepackage{multirow}
\usepackage{geometry}
\usepackage{float}
\usepackage{subfig}

\title{Efficacy of machine-generated annotations}

\author{Samaksh Gulati, Anshit Verma, Manoj Parmar, Palash Chaudhary 
\thanks{Code at - \url{https://github.com/samaksh97/Deep-Learning-Project-}} \\
\texttt{\{sgulati, averma373, mparmar, palash.choudhary\}@gatech.edu} \\
}

\iclrfinalcopy 
\begin{document}

\maketitle

\begin{abstract}

Large “instruction-tuned” language models (i.e., finetuned to respond to instructions) have demonstrated a remarkable ability to generalize zero-shot to new tasks. Nevertheless, they depend heavily on human-written instruction data that is often limited in quantity, diversity, and creativity, therefore hindering the generality of the tuned model. We conducted a quantitative study to figure out the efficacy of machine generated annotations, where we compare the results of a fine-tuned BERT model with human v/s machine-generated annotations. Applying our methods to the vanilla GPT-3 model, we saw that machine-generated annotations were 78.54\% correct and the fine-tuned model achieved a 96.01\% model performance compared to the performance with human labelled annotations. This result shows that machine-generated annotations are an resource and cost effective way to fine-tune down-stream models.

\end{abstract}

\section{Introduction}




The recent NLP literature has witnessed a tremendous amount of activity in building models that can follow natural language instructions \cite{mishra2022crosstask, sanh2022multitask,gupta2023instruction}. These have shown that fine-tuning language models on a collection of data sets described via instructions—substantially improves zero-shot performance on unseen tasks\cite{wei2021finetuned,gupta2021context,scaria2023instructabsa}.
These developments are powered by two key components: large pre-trained language models (LM) and human-written instruction data (e.g., PROMPTSOURCE (\cite{bach2022promptsource}) and SUPERNATURALINSTRUCTIONS (\cite{wang2022supernaturalinstructions}, SUPERNI for short)).
However, collecting such instruction data is costly and often suffers limited diversity given that most human generations tend to be popular NLP tasks, falling short of covering a true variety of tasks and different ways to describe them. Continuing to improve the quality and coverage of instruction-tuned models necessitates the development of alternative approaches for supervising the instruction tuning process \cite{gupta2023targen}.

The objective of the paper is to distil Instruction Fine-Tuning Data from Closed-Source Models to reduce resource overhead and determine how good these models are, in generating annotations for our chosen task. We aim to tackle the task of categorizing textual data into predefined categories or labels. To do this, we come up with three different methodologies. First, we use GPT-3 with zero-shot training. Secondly, we Fine-tune BERT \cite{devlin2019bert}  with human-annotated data. Finally, we fine-tune the BERT model \cite{devlin2019bert} with GPT annotated data to general categories. In this paper we compared the performance of a fine-tuned BERT model with human and machine generated instructions on these three methods. We tried the following experiments throughout our project:
\begin{itemize}
    \item We generated synthetic labels for two different tasks - Classification and Question and Answering. The two task helped us evaluate the correctness of labels generated and performance of fine-tune model holistically as there cover the complexities and semantic logic of English language.
    \item We also implemented 3 different methods to generate labels as we wanted to compare the quality of generated labels on varying involvement of human input.
\end{itemize}

We hope that this project can serve as a reference for quantization of the how good the performance of a fine-tuned model is on machine-generated annotations. We believe that we would also be able to contribute to the cost dynamics management which is a major part of fine-tuning large language models.


\subsection{Datasets}

\subsubsection{Conference Title Classification Dataset}
The study presents a dataset comprising 2,507 research paper titles manually classified into 5 categories, representing different conferences. With an average title length of 9 words, the dataset exhibits a class imbalance, with the majority class comprising 34.5\% of records. We explore the application of stratified sampling to address the class imbalance and enhance the performance of a classification model.

\begin{figure}[h]
    \begin{center}
    \includegraphics[scale=0.5]{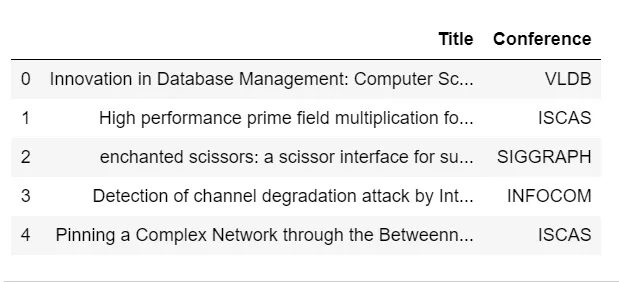}
    \end{center}
    \caption{Dataset Description.}
\end{figure}

\subsubsection{SQuAD Dataset}
 The Stanford Question Answering Dataset (SQuAD) \cite{rajpurkar2016squad} stands out with its extensive set of 107,785 question-answer pairs sourced from 536 Wikipedia articles in SQuAD 1.1. Within these chosen articles, they extracted 23,215 individual paragraphs, ensuring the exclusion of overly small paragraphs. The dataset was then partitioned by articles, allocating 80\% for training, 10\% for development, and 10\% for testing, maintaining a balanced and representative distribution across sets.

 \begin{figure} [h]
     \centering
     \includegraphics[width=0.5\linewidth]{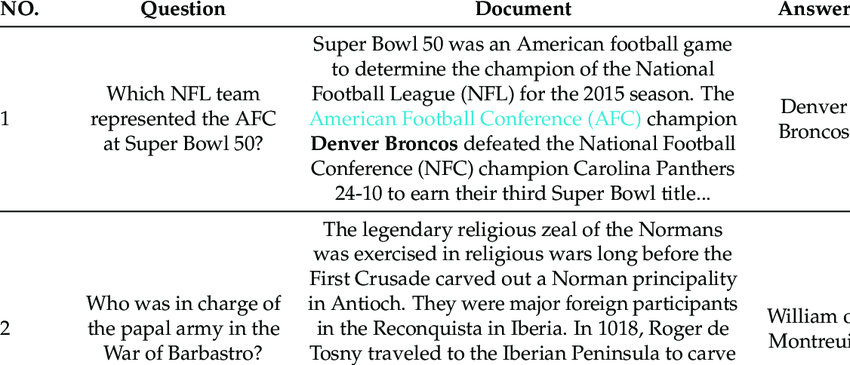}
     \caption{SQuAD description}
     \label{fig:enter-label}
 \end{figure}

\subsection{Fine Tuning Language Models}
Fine-tuning large language models is a process where a pre-trained model, initially trained on a vast and diverse dataset, is further trained (or "fine-tuned") on a smaller, more specific dataset. This process adapts the model to perform better on tasks related to the characteristics of the smaller dataset. The pre-trained model is then trained further on a more specialized dataset. This dataset is usually much smaller than the original training set and focused on a specific domain or task. During fine-tuning, the model's parameters are adjusted to better align with the specifics of the new data. This includes learning task-specific features and nuances. By fine-tuning, the model often achieves higher accuracy and better performance on tasks closely related to the fine-tuning dataset. \\

\begin{figure}[h]
    \centering
    \includegraphics[width=0.6\linewidth]{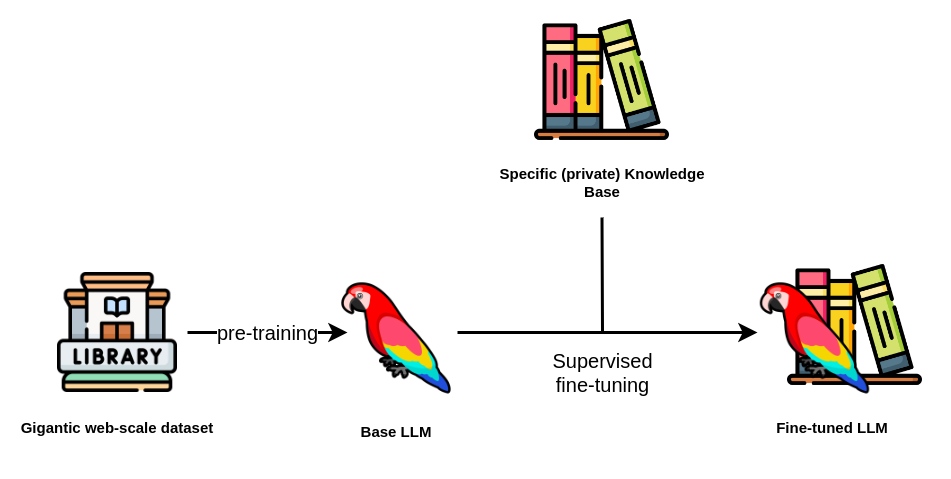}
    \caption{Fine-Tuning LLM}
    \label{fig:enter-label}
\end{figure}

While there are strengths, fine-tuning also has limitations - there's a risk that the model may become too specialized to the fine-tuning dataset and lose its ability to generalize. \\
In summary, fine-tuning large language models is a critical step in tailoring these models for specific tasks or domains, enhancing their performance while mitigating the need for extensive resources required for training large models from scratch.

\section{Related Works}

Fine-tuning language models require collecting a small but specific dataset, in order to adapt and better align the model to a given task. In past works, this data has been manually collected by humans which result in an overhead resource cost in terms of humans and money. Like \cite{wei2021finetuned}, which transformed existing datasets from the research community into an instructional format and aggregated 62 text datasets that are publicly available on Tensorflow Datasets, including both language understanding and language generation tasks, into a single mixture as created an instruction instruction tuning dataset with many tasks from scratch would be resource-intensive.
Other limitations which come from previous methods \cite{wang2023selfinstruct}, which aims to enhance the instruction-following capabilities of pre-trained language models (LMs) using self-generated instructions, is that only 58\% of the outputs were found to be correct and acceptable responses to the instructions, indicating a need for further refinement. 
Even state-of-the-art models like Open AI's InstructGPT\cite{ouyang2022training}, was fine-tune on data which was more aligned to human values. They hired a team of 40 contractors to label the data and collected a dataset of human-written demonstrations of the desired output behavior and some labeler-written prompts.
The limitations as not limited to data collection, ensuring the quality and relevance of the data for the specific task at hand is also crucial. Poor quality data, including inaccuracies, inconsistencies, or irrelevant information, can lead to suboptimal model performance.

\section{Data Collection}
We first started our data collection with widely available datasets - Conference title and SQuAD dataset. For our study we are limiting the size of dataset to 2500 entries for each dataset. This number translates to 2500 question for SQuAD dataset.
Second, we generated synthetic data using GPT-3 \cite{NEURIPS2020_1457c0d6}, where we provided instructions to the agent depending on the task.
\begin{figure}[h]%
    \centering
    \subfloat[\centering SQuAD Dataset]{{\includegraphics[width=0.8\linewidth]{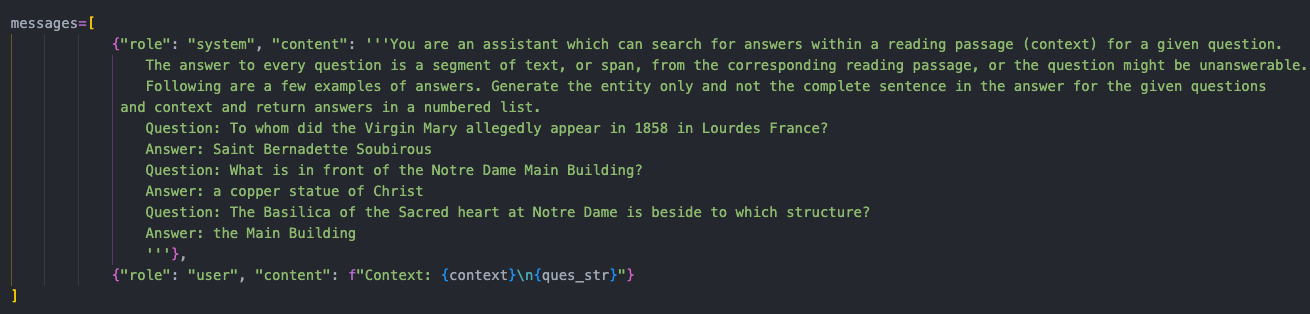} }}%
    \qquad
    \subfloat[\centering Conference Title Dataset]{{\includegraphics[width=0.8\linewidth]{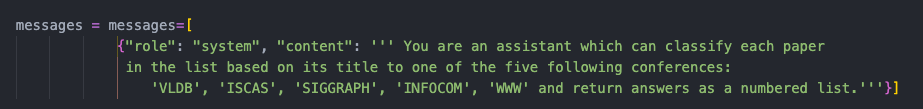}}}%
    \caption{Instructions for synthetic data generation}%
    \label{fig:example}%
\end{figure}

Separate data was generated for each methodology, with and without few shots. Fig. 4(a) shows instructions for SQuAD data generation with few shots, while Fig. 4(b) shows instructions for Conference Title data generation without few shots. For the classification task we also gave options to promote GPT-3 to give answers within those answers so as to decrease chances of hallucination.


\section{Methodology/Experimental Setup}
\subsection{Classification via GPT-3.5}

In this method, we will generate labels using GPT-3.5 with zero shot training and generate synthetic labels for the dataset. The idea here is to evaluate the performance of GPT-3 on creating synthetic labels based on its pre-training corpus and only instruction tuning and no fine-tuning or even few-shots prompting. We will evaluate the efficacy of generated labels by calculating classification accuracy against the actual labels in the dataset.

\begin{figure}[h]
    \begin{center}
    \includegraphics[width=\textwidth]{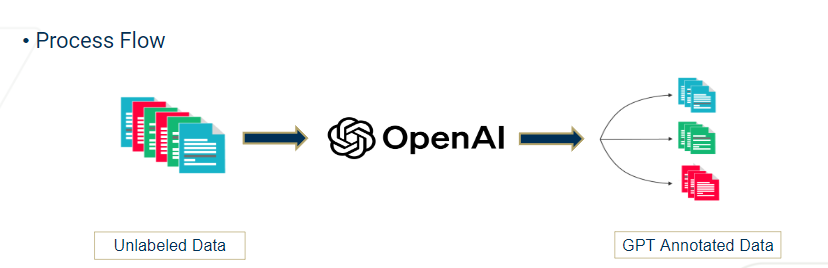}
    \end{center}
    \caption{Classification using zero shot GPT annotated data without fine-tuning.}
\end{figure}

\subsection{Fine-tuning BERT with human annotated data}

In this method, we are fine-tuning pre-trained BERT model with human annotated labeled data. The idea here is to evaluate the performance of a fined-tuned model on actual labeled data and see the upliftment in accuracy for validation data. We will evaluate the performace of the fine-tuned model by calculation classification accuracy on validation data against the actual labels in the dataset. Fig. 6 shows the process we followed for this methodology.

\begin{figure}[h]
    \begin{center}
    \includegraphics[width=\textwidth]{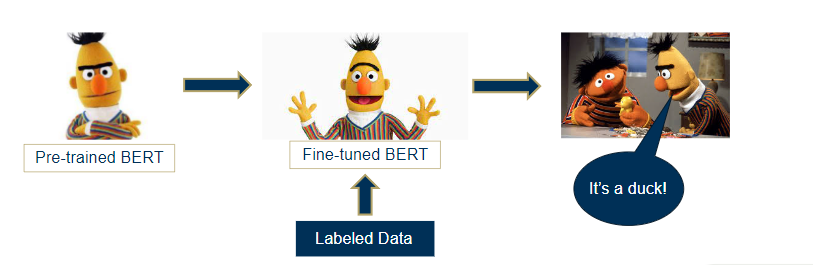}
    \end{center}
    \caption{Fine-tuning BERT with human annotated data.}
\end{figure}

\subsection{Fine-tuning bert with gpt 3.5 annotated data}

This approach combines elements from both methodologies mentioned earlier. Initially, synthetic labels generated through Methodology 1 with instruction tuning are utilized. Subsequently, a pre-trained BERT model is fine-tuned using these synthetic labels or data annotated by GPT. The goal is to assess the performance of a fine-tuned BERT model on GPT-annotated data, comparing its efficacy against actual labels. This proves crucial in scenarios where human-annotated labels are not feasible, and machine annotations via GPT are leveraged to create synthetic labels for fine-tuning. The approach aims to evaluate the model's adaptability to GPT-generated annotations in the absence of human-labeled data. Fig. 7 shows the process we followed for this methodology.

\begin{figure}[h]
    \begin{center}
    \includegraphics[width=1\textwidth]{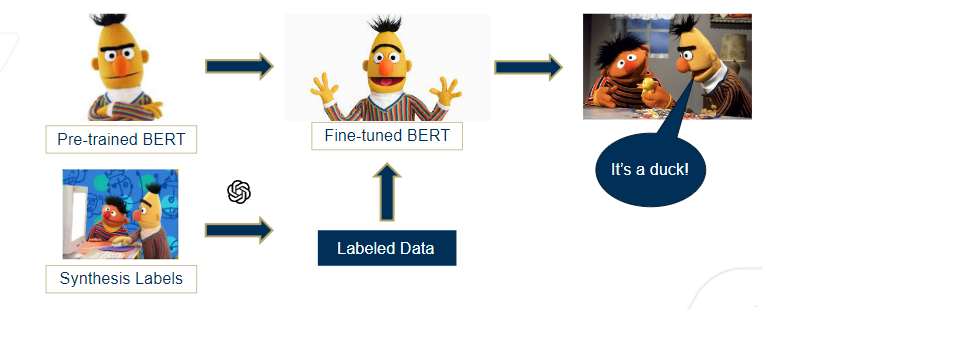}
    \end{center}
    \caption{Fine-tuning BERT with GPT annotated data.}
\end{figure}

\section{Results}

We will compare the results for the above three methodologies on two datasets - Conference Title Data and SQuAD dataset. Intuitively, Methodology 2 should have the best results since the fine-tuning is done on actual labels followed by Methodology 3 in which we are fine-tuning on machine generated labels using GPT. Methodology 1 will have the worst results since we are generating labels using GPT without fine-tuning and zero shot prompting. We are comparing the results on these tasks in the results table \ref{tab:conf_results}

The below table \ref{tab:parameters} represents the parameters used while generating synthetic labels and fine-tuning the BERT model. 

\begin{table}[h]
\centering
\begin{tabular}{|c|c|c|}
\hline
\textbf{Type} & \textbf{Parameters} & \textbf{Value} \\
\hline
API Call & GPT Model Name & gpt-3.5-turbo \\
\hline
\multirow{5}{*}{Fine Tuning} & Model Name & bert-base-uncased \\
& Optimizer & AdamW \\
& Learning Rate & 1e-5 \\
& Epsilon & 1e-8 \\
& Epochs & 5 \\
\hline
\end{tabular}
\caption{Fine Tuning Parameters}
\label{tab:parameters}
\end{table}

\subsection{Conference Title Data}

\begin{table}[ht]
\centering
\resizebox{\columnwidth}{!}{%
\small
\begin{tabular}{cc|ccc|}
\cline{3-5}
\multicolumn{1}{l}{}           & \multicolumn{1}{l|}{} & \multicolumn{3}{c|}{\textbf{Correctly Predicted}}         \\ \hline
\multicolumn{1}{|c|}{\textbf{Category}} &
  \textbf{Total Count} &
  \multicolumn{1}{c|}{\textbf{Methodology 1}} &
  \multicolumn{1}{c|}{\textbf{Methodology 2}} &
  \textbf{Methodology 3} \\ \hline
\multicolumn{1}{|c|}{VLDB}     & 63                    & \multicolumn{1}{c|}{52}  & \multicolumn{1}{c|}{43}  & 50  \\ \hline
\multicolumn{1}{|c|}{ISCAS}    & 130                   & \multicolumn{1}{c|}{106} & \multicolumn{1}{c|}{123} & 109 \\ \hline
\multicolumn{1}{|c|}{SIGGRAPH} & 48                    & \multicolumn{1}{c|}{44}  & \multicolumn{1}{c|}{39}  & 41  \\ \hline
\multicolumn{1}{|c|}{INFOCOM}  & 77                    & \multicolumn{1}{c|}{57}  & \multicolumn{1}{c|}{59}  & 59  \\ \hline
\multicolumn{1}{|c|}{WWW}      & 56                    & \multicolumn{1}{c|}{33}  & \multicolumn{1}{c|}{44}  & 37  \\ \hline
\multicolumn{1}{|c|}{\textbf{F1-score}} &
  \textbf{-} &
  \multicolumn{1}{c|}{\textbf{78.25\%}} &
  \multicolumn{1}{c|}{\textbf{82.8\%}} &
  \textbf{79.5\%} \\ \hline
\end{tabular}%
}
\caption{Conference Title Data Results}
\label{tab:conf_results}
\end{table}

From the table \ref{tab:conf_results}, we can see that Methodology 1 has a weighted average f1-score of 78.25\%. Methodology 2 has the highest f1-score of 82.8\% which is what we expected intuitively. Methodology 3 has an f1-score if 79.5\% which is a slight upliftment from the Methodolgy 1. Fine-tuning on synthetic labels (Methodology 3) vs fine-tuning on actual labels (Methodology 2) only have a 3.3\% absolute difference.

\begin{table}[h]
\centering
\begin{tabular}{|c|c|c|}
\hline
\textbf{Metric Comparison} & \textbf{Human} & \textbf{GPT} \\
\hline
Label Accuracy & 100\% & 78.54\% \\ \hline
Model Performance & 100\% & 96.01\% \\ \hline
Cost & 100\% & 0.12\% \\
\hline
\end{tabular}
\caption{Comparison of Metrics}
\label{tab:metric_comparison}
\end{table}

The above table \ref{tab:metric_comparison} compares the results of human annotated data vs GPT annotated data on a relative scale. We can see that the cost for creating synthetic labels is negligible ,i.e. 0.12\% of the cost of generating human annotated labels via MTurk or other surveys. However, the model performance is not depreciating a lot. It is 96.01\% of what we would expect with a human annotated label.

\subsection{Squad Data}

From the table \ref{tab:squad_results}, we can see that Methodology 1 achieves an accuracy of 39.54\%. Methodology 2, i.e. finetuning BERT with Squad data sample, yields the best results with an accuracy of 48.99\%. Methodology 3 yields the least favourable result with an accuracy of 33.62\%. This is mainly because evaluating subjective question answers on the basis of accuracy might not yield the best results. 

\begin{table}[h]
\centering
\begin{tabular}{|c|c|c|c|}
\hline
\textbf{Dataset} & \textbf{Methodology 1} & \textbf{Methodology 2} & \textbf{Methodology 3} \\
\hline
Squad data sample & 39.54\% & 48.99\% & 33.62\% \\
\hline
\end{tabular}
\caption{Squad Dataset Results}
\label{tab:squad_results}
\end{table}

\begin{figure}[h]
    \begin{center}
    \includegraphics[width=0.5\textwidth]{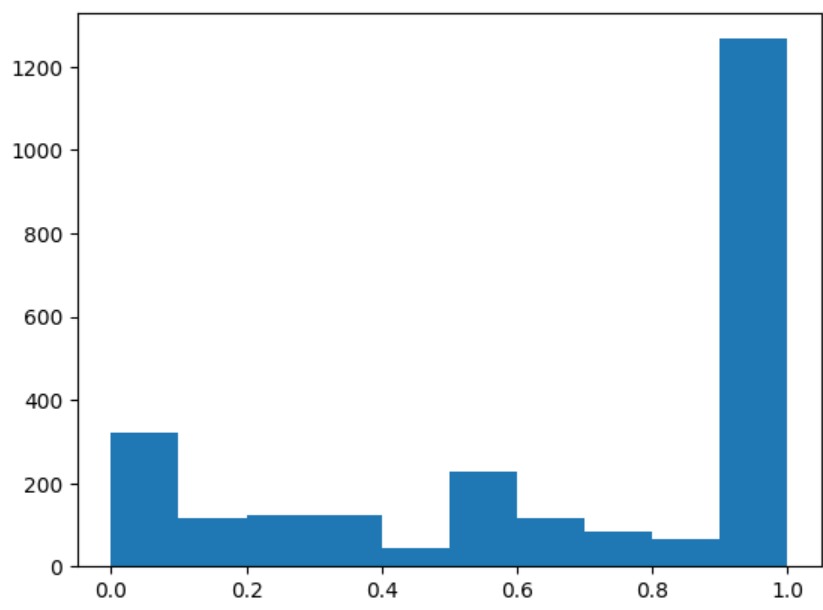}
    \end{center}
    \caption{Squad training data BLEU score distribution}
    \label{fig: bleu_score_dist}
\end{figure}

On evaluating the GPT-generated training data labels, we observed that a lot of incorrect answers were because of one additional word or a different format for the output. Thus, we evaluated the quality of synthetic data using the BLEU score. Using the BLEU score distribution for training sample, as shown in \ref{fig: bleu_score_dist}, we obtained an accuracy of 70.80\% by defining a threshold of BLEU score = 0.5.

For methodology 2, with Squad data sample of 2500 data points, we fine-tuned the BERT model with human-annotated labels. We trained the models for 4 epochs and achieved a validation accuracy of 48.99\%, as shown in \ref{tab:squad_m2_finetuning}.

\begin{table}[h]
\centering
\begin{tabular}{|c|c|c|c|}
\hline
\textbf{Epochs} & \textbf{Train Accuracy} & \textbf{Validation Accuracy}  \\
\hline
1 & 20.99\% & 40.52\% \\ \hline
2 & 60.60\% & 47.68\% \\ \hline
3 & 75.73\% & 48.61\% \\ \hline
4 & 83.55\% & 48.99\% \\ 
\hline
\end{tabular}
\caption{BERT Finetuning using Squad Data (human annotated labels)}
\label{tab:squad_m2_finetuning}
\end{table}

For methodology 3, with Squad data sample of 2366 data points, we fine-tuned the BERT model with GPT-annotated labels. We trained the models for 4 epochs and achieved a validation accuracy of 33.62\%, as shown in \ref{tab:squad_m2_finetuning}.

\begin{table}[h!]
\centering
\begin{tabular}{|c|c|c|c|}
\hline
\textbf{Epochs} & \textbf{Train Accuracy} & \textbf{Validation Accuracy}  \\
\hline
1 & 22.75\% & 25.57\% \\ \hline
2 & 51.83\% & 35.18\% \\ \hline
3 & 68.21\% & 38.52\% \\ \hline
4 & 79.22\% & 33.62\% \\ 
\hline
\end{tabular}
\caption{BERT Finetuning using Squad Data (GPT-annotated labels)}
\label{tab:squad_m3_finetuning}
\end{table}

\section{Conclusion}

During our experiments, we observed that GPT-annotated fine-tuning produces equivalent results to human annotated fine-tuning in the multi-class classification task( 79.5\% for GPT-annotated as compared to 82.8\% for human annotated), whereas the performance deteriorates drastically in the question answering task (33.6\% accuracy for GPT-annotated as compared to 48.99\% for human-annotated). We believe that the difference in performance is because of the difference in the objective tasks. The question-answering task is complex as compared to the classification task and requires a deeper understanding of the training data during fine-tuning. Thus, it is important that training data during fine-tuning is as accurate as possible. Further, the overall performance of the question-answering task is below par with the performance of the classification task. This might be because of limited data and a smaller number of epochs in training.

Future work can include experimenting with larger data, training for higher number of epochs, experimenting with prompts for generating more accurate synthetic response, and identifying better evaluation metric for subjective question-answering.

\bibliographystyle{iclr2021_conference}
\bibliography{iclr2021_conference}


\end{document}